%% file: iclr2026_conference.tex
\documentclass{article} 
\usepackage{iclr2026_conference, times}

\input{math_commands.tex}

\usepackage{hyperref}
\usepackage{url}
\usepackage{graphicx}%
\usepackage{subcaption} 
\usepackage{multirow}%
\usepackage{tcolorbox}%
\tcbset{colback=blue!5!white, colframe=blue!40!black,
        boxrule=0.5pt, arc=4pt, auto outer arc,
        boxsep=5pt, left=8pt, right=8pt, top=8pt, bottom=8pt}

\usepackage{amsmath,amssymb,amsfonts}%
\usepackage{mdframed}
\usepackage{xcolor}

\newmdenv[
  backgroundcolor=gray!10,
  linecolor=gray,
  linewidth=0.5pt,
  roundcorner=5pt,
  innerleftmargin=10pt,
  innerrightmargin=10pt,
  innertopmargin=8pt,
  innerbottommargin=8pt
]{examplebox}
\usepackage{amsthm}%
\usepackage{mathrsfs}%
\usepackage[title]{appendix}%
\usepackage{svg}%
\usepackage{xcolor}%
\usepackage{textcomp}%
\usepackage{manyfoot}%
\usepackage{booktabs}%
\usepackage{algorithm}%
\usepackage{algorithmicx}%
\usepackage{algpseudocode}%
\usepackage{listings}%
\usepackage{soul}
\usepackage{svg}
\usepackage{amssymb}    

\usepackage{pifont}     

\title{ForTIFAI: Fending Off Recursive Training \\ Induced Failure for AI Model Collapse}

\author{
  Soheil Zibakhsh Shabgahi\footnotemark[2] \textsuperscript{1} \quad
  Pedram Aghazadeh\footnotemark[2] \textsuperscript{1} \quad
  Azalia Mirhoseini\textsuperscript{2} \quad
  Farinaz Koushanfar\textsuperscript{1} \\
  \textsuperscript{1} University of California San Diego \\
  \textsuperscript{2} Stanford University \\
  \texttt{\{szibakhshshabgahi,paghazadeh,farinaz\}@ucsd.edu} \quad
  \texttt{azalia@stanford.edu}
}

\newcommand{\llama}[1]{LLaMA-3.2-1B}
\newcommand{\gemma}[1]{Gemma-3-1b-pt}
\iclrfinalcopy 

\begin{document}
\renewcommand{\thefootnote}{\fnsymbol{footnote}}
\footnotetext[2]{Equal contribution.}
\renewcommand{\thefootnote}{\arabic{footnote}}

\maketitle

\begin{abstract}
The increasing reliance on generative AI models is rapidly increasing the volume of synthetic data, with some projections suggesting that most available new data for training could be machine-generated by 2030 \cite{gartner2022synthetic}. This shift to a mainly synthetic content presents a critical challenge: repeated training in synthetic data leads to a phenomenon known as model collapse, where model performance degrades over generations of training, eventually rendering the models ineffective. While the causes of model collapse are increasingly understood, effective mitigation strategies remain scarce. We address this challenge by leveraging a key insight: auto-regressive models tend to generate text sequences to which they assign high confidence (i.e., high log-likelihood).
Based on this observation, we introduce the Truncated-Cross-Entropy (TCE) loss function. TCE mitigates collapse by selectively ignoring high-confidence tokens during training, effectively filtering out likely machine-generated artifacts from the learning process.
Our experiments demonstrate that models trained with TCE not only learn effectively but also exhibit significantly increased resilience, tolerating over 2.3× more synthetic data before the onset of collapse.
In addition, we provide an open-source benchmark for collapse dynamics in mixed-data settings. Our results demonstrate that confidence-aware training objectives can substantially delay collapse onset, offering a practical and generalizable tool for model robustness under synthetic-data exposure.
\end{abstract}

\section{Introduction}\label{sec1}
Generative models have become the foundation for modern AI applications in several modalities, including text, image, code, and audio. Large Language Models (LLMs) such as ChatGPT~\citep{openai2024gpt4technicalreport}, LLaMA~\citep{grattafiori2024llama3herdmodels} and Gemma~\citep{gemmateam2025gemma3technicalreport}, as well as image generators  DALL-E~\citep{ramesh2021zeroshottexttoimagegeneration} and Imagen~\citep{saharia2022photorealistictexttoimagediffusionmodels}, all rely on large datasets scraped from the Web. As these models are continuously updated to reflect recent knowledge and linguistic patterns, the need for ever larger and frequently refreshed training corpora has grown substantially. However, this demand is colliding with a shift in the data landscape: synthetic content is increasingly populating the Internet, contaminating the very datasets used for model training.

This shift raises fundamental concerns. Generative models trained on outputs from earlier generations can degrade over time, a phenomenon known as \textit{model collapse}~\citep{Model_collapse, shumailov2024curserecursiontraininggenerated, dohmatob2024taletailsmodelcollapse}. Even moderate levels of synthetic data (as little as 1\%) have been shown to cause collapse, and scaling the model or the size of the dataset fails to reliably prevent it~\citep{dohmatob2024strongmodelcollapse}. Importantly, this is not a hypothetical scenario: recent analysis of \citet{alemohammad2023selfconsuminggenerativemodelsmad} on the LAION-5B dataset, a cornerstone for many generative models, revealed measurable contamination from synthetic sources.\footnote{Following prior works~\citep{Model_collapse, shumailov2024curserecursiontraininggenerated}, we use the term \textit{contamination} or \textit{poisoning} to describe the recursive use of synthetic data in training, which leads to degradation in performance over generations.} As synthetic data becomes dominant, safeguarding generative model stability becomes an urgent challenge.

Several approaches have been explored to mitigate model collapse. Some studies combine synthetic and real data to slow degradation~\citep{gerstgrasser2024modelcollapseinevitablebreaking}; others rely on post hoc supervision, such as SIMS~\citep{alemohammad2024selfimprovingdiffusionmodelssynthetic}, using labeled synthetic data as negative signal to improve generations or explore decoding strategies~\citep{drayson2025machinegeneratedtextdetectionprevents}. In practice, however, distinguishing real from synthetic data remains a major challenge. Watermarking methods~\citep{zhang2024emmarkrobustwatermarksip, kirchenbauer2024watermarklargelanguagemodels} have been proposed to help identify model-generated content, but they are not widely adopted. Moreover, recent critiques~\citep{schaeffer2025positionmodelcollapsedoes} highlight that many prior studies simulate collapse in unrealistic settings, such as using purely synthetic training data, limiting the applicability of their conclusions. Despite growing interest in this issue, the role of the loss function, central to how models learn, has received little attention in the context of model collapse.

In this work, we present ForTIFAI, showing that model collapse can be mitigated through principled modifications to training loss. We argue that generative models tend to be overconfident in their synthetic predictions (as depicted in Figure~\ref{self_confidence}), causing feedback loops that degrade performance. This issue is further amplified by approximation errors and finite sampling effects during training, which cause models to concentrate probability mass on a shrinking subset of the true data distribution over generations~\citep{shumailov2024curserecursiontraininggenerated, dohmatob2024taletailsmodelcollapse}. To address this challenge, we introduce Truncated Cross-Entropy (TCE), a loss function that formalizes the intuition of ignoring what the model already ``knows'' too well. TCE operates by setting the loss to zero for any token whose predicted probability exceeds a predefined threshold. This prevents the model from overfitting to its own high-conviction and often low-diversity outputs. As a minimal extension of the standard Cross-Entropy (CE) objective, TCE is model-agnostic, computationally efficient, and easy to implement, making it a practical solution for real-world deployment.

In brief, the contributions of ForTIFAI are as follows:
\begin{itemize}
    \item We establish that a model's confidence, measured by sequence probability, serves as a strong signal for identifying its own generated text within mixed datasets.
    \item We introduce Truncated Cross Entropy (TCE), a confidence-aware loss function that leverages this signal to mitigate model collapse, enabling models to tolerate more than $2.3\times$ greater proportions of synthetic data without degradation.
    \item We propose a comprehensive evaluation framework for model collapse that reflects realistic conditions, including the continual accumulation of new data and the need to measure both retained and newly acquired knowledge.
\end{itemize}

\section{Method}
\subsection{Problem Definition}
Model Collapse is the gradual degradation that occurs when a model’s own outputs pollute the training data of its successors. Each new model then learns from these possibly flawed examples, causing a drift from the true data distribution~\citep{Model_collapse}.

The fundamental cause of collapse is the discrepancy between the true data distribution, $P_{data}$, and the model's learned approximation of it, $P_{\theta}$. Because any realizable model has finite capacity and is trained on finite data, its outputs will contain statistical artifacts not present in the true distribution. Compounding this issue, the decoding process itself introduces a further distributional shift. Common decoding algorithms (e.g., greedy search, beam search) are designed to favor high-probability sequences rather than sampling purely from the model's learned distribution. This bias means that even if a model were a perfect representation of the data ($P_{\theta} = P_{data}$), the generated text would over-represent the modes of the distribution.

When these skewed outputs are used to train a successor model, the imperfections are learned and amplified. This iterative process leads to a progressive distributional drift, causing a catastrophic loss of information, particularly in the tails of the distribution, ultimately making the model unusable.

\begin{figure}
    \centering
    \includegraphics[]{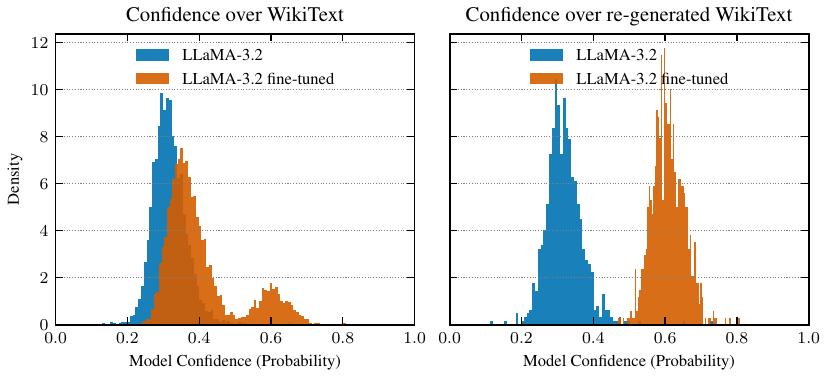}
    \caption{A model exhibits significantly higher confidence in its own generated data compared to unseen data. The histograms show token probabilities from a base \llama{} model (blue) and a fine-tuned version (orange). When evaluating its own synthetic generations (right column), the fine-tuned model shows a clear shift toward higher confidence compared to its evaluation on unseen data (left column). In contrast, the base model's confidence distribution remains consistent across both datasets, identifying elevated confidence as a sign of self-generated text.}
\label{self_confidence}
\end{figure}

\subsection{Key Observation}
\label{sub:key_observation}
Figure~\ref{self_confidence} illustrates a consistent gap in model confidence between self-generated samples and unseen data. We exploit this difference as a signal to identify and remove the effect of synthetic samples during training. This gap is a side effect of the approximation error and the greedy sampling performed for generating text \citep{drayson2025machinegeneratedtextdetectionprevents}. 

\subsection{Truncated Cross Entropy (TCE)}
Building on the findings from Section~\ref{sub:key_observation}, we introduce a novel loss function, Truncated Cross Entropy (TCE). In TCE, confident predictions are explicitly masked to remove their influence on training. By selectively masking overly confident predictions, TCE forces the model to learn from less-certain datapoints, thereby preserving distribution tails compared to CE. We formulate TCE as:

$$TCE(p_t)=\chi_\gamma(p_t)\times CE(p_t)$$
$$\chi_\gamma(p_t)=\begin{cases}
    1 \textit{ if } p_t\leq\gamma \\
    0 \textit{ if } p_t>\gamma
\end{cases}$$

Here, $p_t$ is the probability of the correct class, $\chi_\gamma(.)$ is an indicator function, $CE(.)$ is the cross-entropy loss, and $\gamma \in [0, 1]$ is the confidence threshold. Setting $\gamma$ close to 1 recovers standard CE loss, while smaller values selectively suppress high-confidence predictions. In Appendix~\ref{app:focal} we examine Focal Loss as another alternative confidence-aware loss function with similar properties.

\subsection{Mathematical Insight}
The growing reliance on synthetic data in generative model training pipelines has given rise to a self-consuming feedback loop wherein models are recursively trained on their own outputs. 

\begin{figure}[t]
\centering
\includegraphics[width=0.5\textwidth]{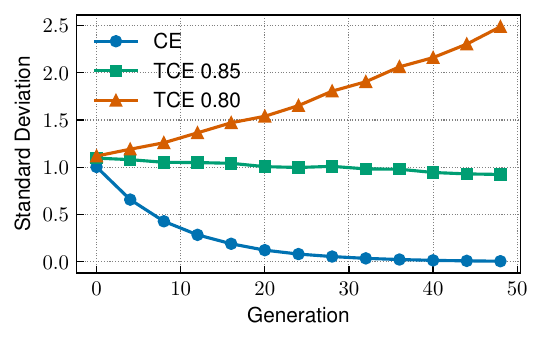}
\caption{Effect of TCE on a one-dimensional Gaussian estimator under a fully synthetic training loop. In each generation, $10,000$ samples are generated with sampling bias $\gamma=0.9$. Standard estimation with Cross Entropy (CE) leads to rapid variance collapse. By contrast, TCE with a well-chosen threshold ($\gamma=0.85$) delays collapse substantially. Conversely, if the threshold is set too small ($\gamma=0.80$), the variance may diverge.}
\label{fig:intuition}
\end{figure}

In a fully synthetic training loop, where each model \( G_t \) is trained solely on samples from \( G_{t-1} \), \citet{alemohammad2023selfconsuminggenerativemodelsmad} prove that the model's estimated covariance \( \sigma_t \) collapses almost surely:

\begin{equation}
    \mathbb{E}[\sigma_t \mid \sigma_{t-1}] = \lambda \sigma_{t-1}, \quad \text{with } \lambda \leq 1, \qquad \Rightarrow \quad \sigma_t \xrightarrow{a.s.} 0.
\end{equation}

where $\lambda$ represents sampling bias, contracting the variance at each iteration, leading to rapid loss of sample diversity, as illustrated in Figure~\ref{fig:intuition}. This decrease in variance is empirically observed across models such as DDPMs~\citep{ho2020denoisingdiffusionprobabilisticmodels} and StyleGAN-2~\citep{karras2020analyzingimprovingimagequality}.

To build intuition on how to counteract this collapse, we analyze a one-dimensional Gaussian model where data is drawn from \( X_t \sim \mathcal{N}(0, \sigma_t^2) \). Our strategy is to modify the training objective to focus on low-probability samples. In our Gaussian case, this is equivalent to training only on samples from the tails of the distribution, i.e., where \( |X_t| \geq a \sigma_t \) for a chosen threshold \( a > 0 \).

The conditional variance of this truncated distribution is given by:
\begin{equation}
\mathrm{Var}(X_t \mid |X_t| \geq a \sigma_t) = \eta(a) \cdot \sigma_t^2,
\end{equation}
where \( \eta(a) \) is a variance amplification factor. Using known properties of the truncated normal distribution, this factor is calculated as:
\begin{equation}
\eta(a) = 1 + \frac{a\,\phi(a)}{1-\Phi(a)}
\end{equation}
where \( \phi(\cdot) \) and \( \Phi(\cdot) \) are the standard normal PDF and CDF, respectively. Crucially, \( \eta(a) > 1 \) for all \( a > 0 \).

By incorporating this variance amplification into the self-consuming loop, the recurrence for the variance becomes:
\begin{equation}
\mathbb{E}[\sigma_{t+1}^2] = \lambda \cdot \eta(a) \cdot \mathbb{E}[\sigma_t^2].
\end{equation}
The collapse can now be slowed by choosing a threshold \( a \) such that the combined factor \( \lambda \cdot \eta(a) \approx 1 \), stabilizing the training process by preserving sample diversity.

We argue that overconfident predictions on common tokens can dominate the training signal, ultimately causing the model to under-represent less frequent but meaningful patterns. If not addressed, this effect progressively narrows the model’s output distribution, highlighting the importance of suppressing overconfident learning.

By generalizing this principle from a Gaussian model to general generative models, we propose augmenting loss functions to address this issue. \textbf{Truncated Cross Entropy (TCE)} explicitly masks the contribution of high-confidence predictions. This shifts the training signal toward low-confidence, often underrepresented tokens, mitigating the tail-vanishing effect and reducing the recursive amplification of statistical errors highlighted above.

\section{Model Collapse Evaluation Framework}
\label{sec:eval_framework}

To realistically assess the impact of \textit{Model Collapse}, we adopt an experimental framework inspired by~\cite{gerstgrasser2024modelcollapseinevitablebreaking}. Initially, web data was predominantly human-generated and often curated for quality. Over time, however, machine-generated content, often indistinguishable from human-written text, has increasingly supplemented online data. Our framework simulates a realistic model collapse process by progressively adding synthetic data to the training set. We evaluate the methods on math, reasoning, and knowledge recall benchmark performance. Moreover, we analyze the number of synthetic training iterations it takes for the model to lose performance below a certain threshold, which we call the time to failure.

\begin{figure}[t]
    \centering
    \includegraphics[width=1.0\textwidth]{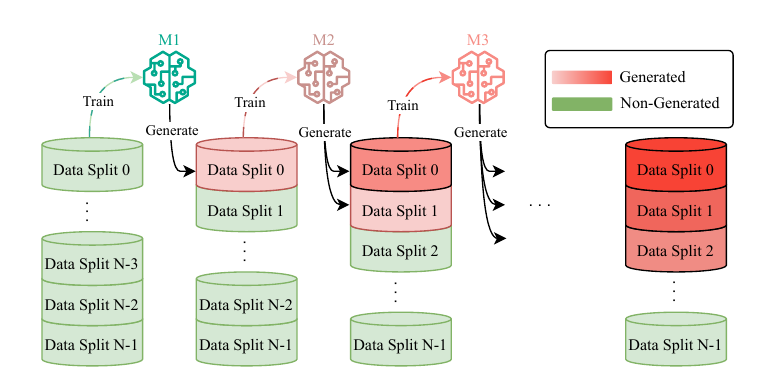}
    \caption{Our experimental setup simulates model collapse, illustrating the transition from predominantly human-generated content to mostly synthetic datasets. $M_i$ denotes the $i$-th generation of recursively trained models. The clean data is initially split into $N$ equally sized splits. At each generation, the entire dataset from the previous iteration is regenerated and used as training data for the next iteration. One split of clean data is added to this training dataset to simulate the accumulation of non-generated data into the training set. 
    The red color for each data split indicates the contamination from recursively training on the self-generated data, resulting in lower quality (darker red color)}
    \label{experiment_structure}
\end{figure}

As illustrated in Figure~\ref{experiment_structure}, the experiment proceeds in multiple stages. At each stage, the dataset simulates a large-scale pretraining corpus, composed of both authentic and self-generated data. At every step, we regenerate the previous stage's dataset using the latest model and concatenate the dataset with a fixed amount of new human-written text. This new dataset is then used to further train the model, as detailed further in Algorithm~\ref{algo1}.

This setup also follows the conditions mentioned in \cite{schaeffer2025positionmodelcollapsedoes}:

\begin{enumerate}
    \item \textbf{Increasing Pre-training Data:} Unlike previous studies with fixed dataset sizes, our setup continuously increases the pre-training data at each generation. This growth mirrors real-world trends among state-of-the-art models; for example, LLaMA 1, 2, and 3 were trained on 1.4, 2, and 15 trillion tokens, respectively. Although dataset growth has eventual limits, these bounds remain extremely high (approximately one quadrillion tokens)~\cite{villalobos2024rundatalimitsllm}, far exceeding current dataset sizes.

    \item \textbf{Synthetic Data Accumulation Alongside Real Data:} Another distinctive feature of our framework is its simultaneous training on both real and synthetic data. Unlike previous work, we do not discount the effect of real unseen data on model collapse. We argue that the real unseen data is the main reason behind training the new generation of models in the first place.
    
    \item \textbf{Decreasing Proportion of Real Data:} A primary concern highlighted in recent literature is the accelerating generation rate of synthetic data and its impact on future generative models. Our experimental setup realistically emulates this crucial aspect (e.g. in our setup with 6 stages, we have 100\%, 50\%, 33\%, 25\%, 20\%, and 15\% real data at each stage, respectively).
\end{enumerate}

\begin{algorithm}
\caption{Recursive data generation with an LLM}\label{algo1}
\begin{algorithmic}[1]
\State \textbf{Inputs:}
\State \hspace{0.5em} $N \geq 2$
\State \hspace{0.5em} $N+1$ splits of authentic dataset $D_0^{(0)},\dots,D_N^{(0)}$, each of equal size

\Comment{$D_i^{(j)}$ is the $j$-th generation of the $i$-th split.}
\State \hspace{0.5em} $N+1$ evaluation sets $Eval_0,\dots,Eval_N$, each of equal size

\Comment{$Eval_i$ is based on the content of $D_i^{(0)}$.}
\State $M_0 \Leftarrow$ Pretrained model
\For{$i=0 \to N$}
    \State Train $M_i$ on $[D_0^{(i)}, D_1^{(i-1)}, \dots, D_i^{(0)}]$
    \For{$k=0 \to i$}
        \State $D_k^{(i+1-k)} \Leftarrow M_i(D_k^{(i-k)})$ 
        \State Evaluate $M_i$ on $Eval_k$
    \EndFor
    \State $M_{i+1} \Leftarrow M_i$
\EndFor
\end{algorithmic}
\end{algorithm}

\section{Results}\label{sec2}
To demonstrate the effectiveness of \texttt{Truncated Cross Entropy}, we evaluate it across three settings: Transformers, Variational Autoencoders, and Gaussian Mixture Models. For Transformers, we focus on language modeling, testing LLaMA 3~\cite{grattafiori2024llama3herdmodels} and Gemma 3~\cite{gemmateam2025gemma3technicalreport} on benchmarks for mathematical reasoning, logical inference, and factual recall. For image generation, we use Variational Autoencoders, and for mixture modeling, we evaluate standard GMMs. Across all experiments, our loss function mitigates model collapse without degrading overall training performance.

\subsection{Experiment Details}
\label{subsec:experiment_setup}
We apply the recursive training setup illustrated in Figure~\ref{experiment_structure} on two datasets: Wikitext-2-raw-v1~\cite{merity2016pointersentinelmixturemodels} and a newly curated dataset called \textit{Imagination of Web}.

\subsubsection{Datasets}
Wikitext-2-raw-v1 is a language modeling dataset derived from high-quality Wikipedia articles. Unlike its tokenized counterpart, it preserves raw formatting—including punctuation, casing, and whitespace—making it well-suited for evaluating a model's ability to handle long-range dependencies and natural language structure.

The second dataset, \textit{Imagination of Web}, is a composite dataset constructed from subsets of Wikitext, HellaSwag~\cite{zellers2019hellaswagmachinereallyfinish}, and GSM8k~\cite{cobbe2021trainingverifierssolvemath}. It is designed to test the generalization of our loss functions across multiple tasks, such as commonsense, math, and factual recall.

\subsubsection{Framework Parameters}
We used the framework described in Section~\ref{sec:eval_framework} to simulate a progressive model collapse scenario.

For our experiments, we set the number of stages to $N = 6$. We found that model collapse was not reliably observable for fewer than six stages, while increasing $N$ beyond six introduced significant computational overhead with diminishing returns under our resource constraints. For simplicity, we used a fixed amount of data at each stage from the original dataset.

\subsubsection{Hyperparameters}
For both LLaMA and Gemma model families, we tuned and used the best-performing hyperparameters. All models were fine-tuned using a learning rate of $2 \times 10^{-5}$, with a batch size of 64. The training was conducted over 10 epochs per stage with the best performing checkpoints saved.

\subsection{Evaluation Benchmarks}
To comprehensively assess the impact of our loss functions on model performance and resilience to model collapse, we perform evaluations on a mix of standard and custom benchmarks. These benchmarks test various capabilities including factual recall, linguistic coherence, commonsense, and mathematical reasoning. In addition to widely used benchmarks like BLiMP, Hellaswag, and GSM8K, we introduce a new question-answering benchmark—\textit{Knowledge Retention test} (KR-test) to measure how well models retain training-set facts.

\subsubsection{Knowledge Retention Test}
Validation perplexity is a common metric for evaluating LLMs, but it primarily reflects predictive fluency; how well a model guesses the next token in unseen text. While useful for assessing grammar and style, it does not quantify factual knowledge retention. Moreover, low perplexity does not guarantee meaningful or human-like output, as humans do not always select the most likely next word.

To address this gap, we introduce the \textit{Knowledge Retention Test} (KR-test), which measures factual \textit{retention} from the training data rather than general fluency.

Given a body of training text, we divide it into complete paragraphs and generate five questions per paragraph. Each question consists of a context, a true continuation, and a false continuation.

We compute the log probability of the model for two sequences: 
\begin{itemize}
    \item \textit{Context + Factually correct completion}
    \item \textit{Context + Factually incorrect completion}
\end{itemize}

When the log probability is higher for the true continuation, the question is considered correctly answered. The KR-test score is the model's total accuracy over all the questions.

A high KR-test score indicates that the model has internalized the factual content of the paragraph. This benchmark specifically tests factual retention from the training data. More details on KR-test can be found in Appendix~\ref{apxKRT}. In Fig.~\ref{accuracy_total_stages} superiority of TCE compared to the baseline is shown after six iterations in correctly answering questions about Wikitext.

\begin{figure}[h]
\centering
\hspace{1cm}
\includegraphics[]{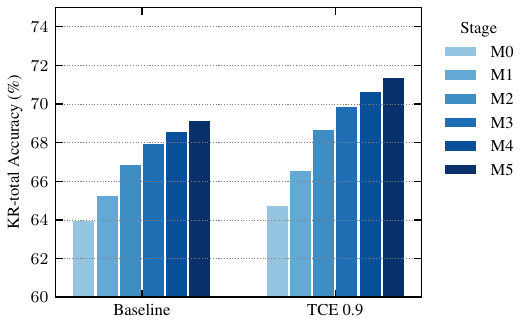}
\caption{Total knowledge retention (KR-total) accuracy with \llama{} model trained on Wikitext. TCE retains learning capability and even exceed the baseline throughout different stages.}\label{accuracy_total_stages}
\end{figure}

\subsubsection{Math, Reasoning, and Coherence}
In addition to factual retention, we evaluate model's performance across three reasoning and language understanding benchmarks. These evaluations assess whether our loss functions preserve or enhance generalization and linguistic structure, both of which are vulnerable under recursive training regimes.

To evaluate the impact of model collapse on reasoning and linguistic capability, we use three standard benchmarks: GSM8K \cite{cobbe2021trainingverifierssolvemath}, Hellaswag \cite{zellers2019hellaswagmachinereallyfinish}, and BLiMP \cite{warstadt2023blimpbenchmarklinguisticminimal}.

\begin{itemize}
    \item \textbf{GSM8K} measures mathematical reasoning by testing the model's ability to solve grade-school level word problems. Success on this benchmark reflects the model’s ability to perform multi-step arithmetic reasoning.

    \item \textbf{Hellaswag} evaluates commonsense reasoning through multiple-choice sentence completion tasks. It requires the model to infer plausible continuations based on everyday scenarios.
    
    \item \textbf{BLiMP} tests grammatical coherence and syntactic structure. It consists of minimal pairs where the model must choose the grammatically correct option. High performance indicates strong control over syntax and language structure.
    
\end{itemize}

Together, these benchmarks assess the model's ability to generalize, reason, and maintain coherence under different forms of linguistic stress. For GSM8K and Hellaswag, we use the Language Model Evaluation Harness~\cite{eval-harness} to ensure consistent evaluation across models.

\subsection{Wikitext dataset}\label{subsec:m5r_results}
To assess performance under idealized, non-collapsing conditions, we also evaluate models trained exclusively on original, non-synthetic data. The results, shown in Table~\ref{tab:m0_llama_gemma_comparison_Wikitext}, confirm that \texttt{TCE 0.9} maintains strong performance in streamline training settings.

Table~\ref{tab:m5_llama_gemma_comparison_Wikitext} presents the results after 5 stages of synthetic data accumulation as discussed in Section~\ref{sec:eval_framework}. The evaluation spans across the full benchmark suite, excluding GSM, since the Wikitext training dataset does not include any math or reasoning. Overall, \texttt{TCE 0.9} outperforms \texttt{CE}, especially in preserving factual knowledge.

\begin{table}[h]
\caption{Comparison of LLaMA and Gemma models on M0 (clean and non-generated data) across Blimp, Hellaswag, GSM8k, and KR-test on \textit{Wikitext dataset}. For LLaMA and Gemma, TCE outperforms baseline (CE) when trained on real non-generated data. The highest value per block is \textbf{bolded}.}
\label{tab:m0_llama_gemma_comparison_Wikitext}
\centering
\begin{tabular}{l|l|c|c|c|c}
\toprule
\textbf{Model} & \textbf{Method} & \textbf{Blimp} & \textbf{Hellaswag} & \textbf{KR-test} & \textbf{Average} \\
\midrule
\multirow{3}{*}{LLaMA}
& Baseline   & 75.1\% & 58.5\% &  64.0\% & 65.86\% \\
& TCE 0.9   & \textbf{76.8\%} & \textbf{61.8\%} & \textbf{64.8\%} &  \textbf{67.8\%}\\
\midrule
\multirow{3}{*}{Gemma}
& Baseline   & 71.2\% & 63.9\% & 67.7\% & 67.6\% \\

& TCE 0.9    & \textbf{73.9\%} & \textbf{64.6\%}  & \textbf{68.2\%} & \textbf{68.9\%}\\
\bottomrule
\end{tabular}
\end{table}

\begin{table}[h]
\caption{Comparison of M5 benchmark results across LLaMA and Gemma models for Baseline, and TCE 0.9 methods on \textit{Wikitext dataset}. The highest value per block is \textbf{bolded}.}
\label{tab:m5_llama_gemma_comparison_Wikitext}
\centering
\begin{tabular}{l|l|c|c|c|c}
\toprule
\textbf{Model} & \textbf{Method} & \textbf{Blimp} & \textbf{Hellaswag} & \textbf{KR-test} & \textbf{Average} \\
\midrule
\multirow{3}{*}{LLaMA}
& Baseline   & 63.3\% & 45.7\% & 69.2\% & 59.4\% \\
& TCE 0.9   & \textbf{63.6\%} & \textbf{48.4\%} & \textbf{71.4\%} & \textbf{61.13\%} \\
\midrule
\multirow{3}{*}{Gemma}
& Baseline   & 61.2\% & 52.4\%  & 68.1\% & 60.56\% \\
& TCE 0.9    & \textbf{61.8\%} & \textbf{53\%} & \textbf{69.7\%} & \textbf{61.5\%} \\
\bottomrule
\end{tabular}
\end{table}

\begin{figure}[t]
    \centering
    \includegraphics[]{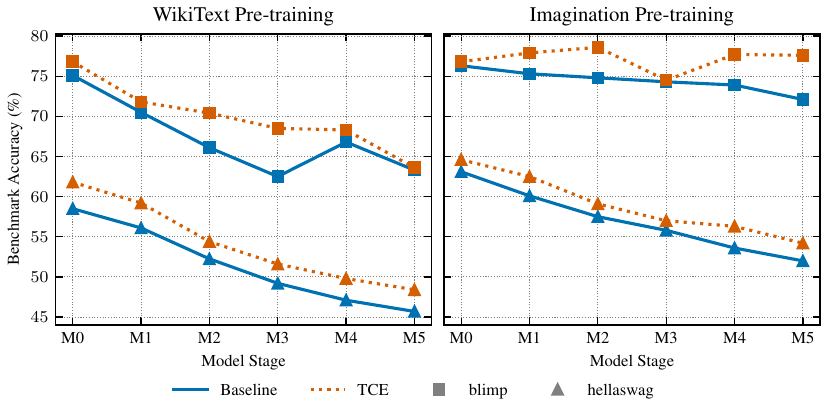}
    \caption{Our proposed solution outperforms baselines (CE) across all benchmarks consistently when experiment is done on both datasets. Figures (a) and (b) show the results of \llama{} across different benchmarks.}\label{accuracy_hellaswag_Wikitext_stages}
\end{figure}

Performance of both models when experiment is conducted on Wikitext dataset is shown in Fig.~\ref{accuracy_hellaswag_Wikitext_stages} where \texttt{TCE} consistently outperforms \textbf{CE} during different stages of the experiment.

\subsection{Imagination-of-Web dataset}\label{subsec:iow_results}
To evaluate in a setting closer to real-world pre-training, we adopt the \textit{Imagination-of-Web} (IoW) corpus—a composite of factual prose (Wikitext), commonsense reasoning (HellaSwag), and mathematical problem solving (GSM8K).

Table~\ref{tab:m0_llama_gemma_comparison} reports clean-data results. Here, \texttt{TCE} surpasses \texttt{CE} by $1.8\%$ on \llama{} and $1.3\%$ on \gemma{}, confirming the strong learning capability of performance under non-collapsing conditions.

\begin{table}[h]
\caption{Comparison of LLaMA and Gemma models on M0 (clean and non-generated data) across Blimp, Hellaswag, GSM8k, and KR-total on \textit{Imagination dataset}. The highest value per block is \textbf{bolded}, and the second highest is \underline{underlined}.}
\label{tab:m0_llama_gemma_comparison}
\centering
\begin{tabular}{l|l|c|c|c|c|c}
\toprule
\textbf{Model} & \textbf{Method} & \textbf{Blimp} & \textbf{Hellaswag} & \textbf{GSM8k} & \textbf{KR-total} & \textbf{Average} \\
\midrule
\multirow{3}{*}{LLaMA}
& Baseline  & 76.3\% & 63.1\% & 8.1\% & 63.6\% & 58.78\% \\
& Focal ($\gamma=2$)   & \textbf{79.6\%} & \textbf{65.2\%} & \textbf{9.5\%} & \textbf{64.8\%} & \textbf{60.62\%} \\
& TCE 0.9  & \underline{76.8\%} & \underline{64.6\%} & \underline{9.1\%} & \underline{64.8\%} & \underline{59.52\%} \\
\midrule
\multirow{3}{*}{Gemma}
& Baseline  & 73.1\% & 65\% & \textbf{6.2\%} & 64.1\% & 52.1\% \\
& Focal ($\gamma=2$)   & \textbf{77.2\%} & \textbf{65.7\%} & 4.5\% & \underline{64.2\%} & \textbf{52.9\%} \\
& TCE 0.9   & \underline{76\%} & 65\% & \underline{5.2\%} & 64\% & \underline{52.55\%} \\
\bottomrule
\end{tabular}
\end{table}

Table~\ref{tab:m5_llama_gemma_comparison} shows outcomes after five rounds of synthetic data accumulation. \texttt{TCE 0.9} outperforms \texttt{CE} by $3.6\%$ on \llama{} and $0.4\%$ on \gemma{}. As on Wikitext, TCE preserves factual and reasoning accuracy, reinforcing its robustness against model collapse in heterogeneous training scenarios.

\begin{table}[h]
\caption{Comparison of M5 benchmark results across LLaMA and Gemma models for Baseline and TCE 0.9 method on \textit{Imagination dataset}. The highest value per block is \textbf{bolded}.}
\label{tab:m5_llama_gemma_comparison}
\centering
\begin{tabular}{l|l|c|c|c|c|c}
\toprule
\textbf{Model} & \textbf{Method} & \textbf{Blimp} & \textbf{Hellaswag} & \textbf{GSM8k} & \textbf{KR-total} & \textbf{Average} \\
\midrule
\multirow{3}{*}{LLaMA}
& Baseline  & 72.1\% & 52.0\% & 5.9\% & 68.7\% & 52.94\% \\
& TCE 0.9  & \textbf{77.6\%} & \textbf{54.2\%} & \textbf{6.2\%} & \textbf{71.0\%} & \textbf{55.52\%} \\
\midrule
\multirow{3}{*}{Gemma}
& Baseline  & 71\% & \textbf{53.4\%} & 3.9\% & 71.6\% & 49.97\% \\

& TCE 0.9   & \textbf{71.7\%} & 53.1\% & \textbf{4.2\%} & \textbf{74.1\%} & \textbf{50.77\%} \\
\bottomrule
\end{tabular}
\end{table}

Performance of both models when experiment is conducted on Imagination dataset is shown in Fig.~\ref{accuracy_hellaswag_Wikitext_stages}, where \texttt{TCE} consistently outperform \texttt{CE} during different stages of the experiment.

\subsection{Time to failure}
\begin{figure}[t]
  \centering
  \includegraphics{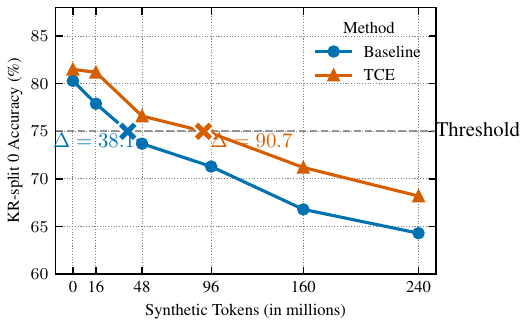}
  \caption{Our proposed loss functions effectively mitigate model collapse in a fully synthetic subset of the dataset. The figure above corresponds to \llama{}'s accuracies on answering questions related to the dataset they were trained on. Red dotted lines provided for better comparison of time to failure. The number of stages until failure is shown as $\Delta$.}
    \label{fig:model_collapse_results}
\end{figure}

We define \textit{time to failure} as the number of re-training tokens it takes a model in the context of this framework to degrade significantly compared to the initial model. Specifically, we analyze the performance of the model on the first partition of the dataset, which is re-generated at each stage. Moreover, since the KR-test on average has a baseline of 50\% and a maximum of 100\% is achievable, we define failure as performance dropping below 75\%, which is exactly half the maximum gain one can expect from training the model. As illustrated in Figure~\ref{fig:model_collapse_results}, Cross Entropy collapses (reaches $75\%$ accuracy) between the first and the second generation and both TCE reaches this point after three stages of self-consuming generation, increasing the time to failure by $2.3\times$.

\subsection{KL Divergence}
To quantify distributional drift across generations, we measured the Kullback–Leibler (KL) divergence between the token distributions of synthetic datasets and the original WikiText-2 corpus. Each dataset is tokenized with the model’s tokenizer, and normalized token frequencies are used to compute the KL divergence between the original and synthetic distributions.

\begin{figure}
    \centering
    \includegraphics[]{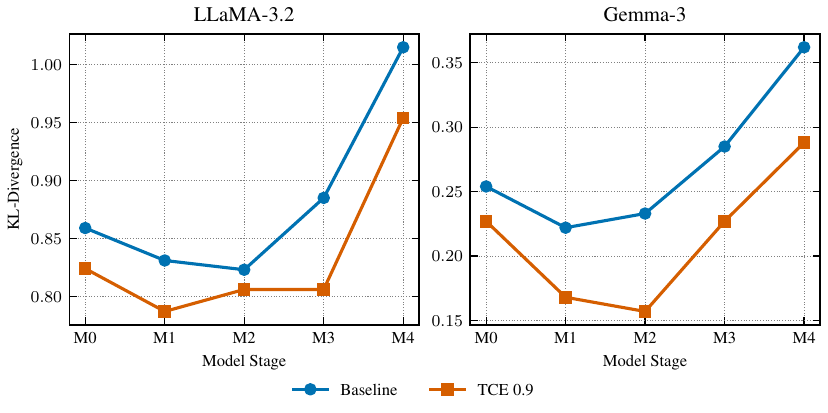}
    \caption{To quantify collapse dynamics, we measured the Kullback-Leibler (KL) divergence between the distribution of model-generated outputs and the original training data distribution over successive generations of the WikiText dataset. Models trained with Cross Entropy showed rapidly increasing KL divergence, indicating distributional drift and loss of diversity. In contrast TCE and significantly slowed the divergence growth, effectively preserving distributional similarity across generations.}
    \label{fig:kl-divergence}
\end{figure}

Figure~\ref{fig:kl-divergence} shows that models trained with TCE maintain a substantially lower KL divergence than those trained with Cross Entropy, indicating better preservation of the original data distribution.

\section{Conclusion}
We have demonstrated that a confidence-aware loss function, Truncated Cross-Entropy (TCE), provides a simple and powerful defense against model collapse. Grounded in a mathematical framework linking overconfidence to distributional drift, TCE extends a model's operational lifetime by over $1.7\times$ compared to standard cross-entropy. Our benchmarks confirm that this approach not only delays collapse but also improves knowledge retention and generalization in environments with accumulating synthetic data, as validated by slower KL divergence growth.

Crucially, the principle of filtering overconfident outputs is not domain-specific. We verified its efficacy in diverse generative systems, including Gaussian Mixture Models and Variational Autoencoders (Appendices~\ref{app:gmm}, \ref{app:vae}). As the line between human and machine-generated data continues to blur, such robust, low-overhead training methods will be essential for building stable and trustworthy AI.

\subsection{Future directions}
We approached model collapse from a training perspective and proposed a class of loss functions that mitigate collapse effectively without sacrificing performance. By reweighting token probabilities to favor underrepresented or underestimated outputs, these losses retain the simplicity of cross-entropy while offering greater resilience during recursive training.

Looking ahead, several research avenues merit further exploration:

\begin{itemize}
    \item \textbf{Broader Model Classes:} Model collapse has been extensively studied in image generation domains. It remains an open question whether similar loss-based strategies can be adapted to diffusion models~\citep{ho2020denoisingdiffusionprobabilisticmodels} and generative adversarial networks (GANs)~\citep{goodfellow2014generativeadversarialnetworks}, where diversity collapse is also a central concern.

    \item \textbf{Heterogeneous Model Ecosystems:} Our setup assumes a single model trained recursively across generations. However, real-world content generation is inherently multi-agent and heterogeneous, involving a mixture of open and closed-source models. Extending our analysis to simulate ecosystems of interacting generative models could provide deeper insights into how synthetic content evolves when shaped by a plurality of architectures and training paradigms.
    
    \item \textbf{Collapse-Aware Generation:} While our approach targets the training process, model collapse can also be addressed during the generation phase. Combining our training-side solution with decoding-aware techniques such as~\citet{drayson2025machinegeneratedtextdetectionprevents} may yield complementary benefits, improving the quality and diversity of generated data further.
\end{itemize}

\bibliography{iclr2026_conference}
\bibliographystyle{iclr2026_conference}

\appendix
\section{Appendix}
\begin{appendices}
\section{Focal Loss}
\label{app:focal}
To complement the main results, we additionally evaluated Focal Loss (FL) in our self-consuming generation framework. While the core paper focuses on Cross Entropy (CE), Truncated Cross Entropy (TCE), and their performance trade-offs, we found it instructive to include FL due to its growing interest in generative modeling. FL shares with TCE the goal of down-weighting confident predictions, which may help preserve the tail of the token distribution during continued generation.

Focal Loss (FL), originally introduced for addressing class imbalance in image segmentation tasks \cite{lin2018focallossdenseobject}, emphasizes hard, misclassified examples. It is defined as:

$$
FL(p_t)=-(1-p_{t})^\gamma\times \log({p_t})
$$

where $p_t$ is the probability of the correct class, and $\gamma$ is a tunable hyperparameter that controls the down-weighting of well-classified samples. FL reduces the loss for confident predictions, acting similarly to cross-entropy on uncertain examples, thus helping avoid overfitting to self-generated tokens.

Recent work \cite{Rege_Cambrin_2024} has shown FL's effectiveness in language modeling, outperforming CE on commonsense reasoning and math tasks. In the context of this work, FL aligns with our objective of mitigating model collapse by emphasizing rare token learning and preserving the tail of the output distribution.

\begin{figure}[h]
\centering
\includegraphics[]{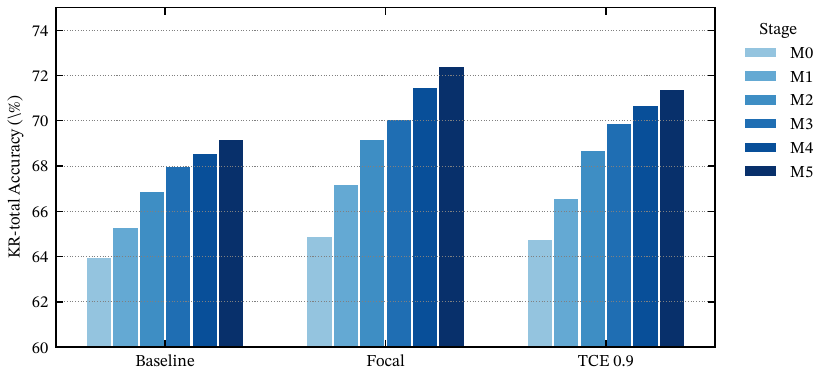}
\caption{Total knowledge retention (KR-total) accuracy with \llama{} model trained on Wikitext. TCE retains learning capability and even exceeds the baseline throughout different stages.}
\label{accuracy_stages}
\end{figure}

\vspace{1mm}
\noindent
Figure~\ref{accuracy_stages} presents the total Knowledge Retention (KR-total) accuracy across generations for \llama{} on WikiText. TCE maintains high performance throughout training, while CE exhibits significant degradation. FL performs comparably to TCE in early stages but begins to degrade slightly earlier, indicating less robustness over long sequences of regeneration.

\begin{figure}[h]
  \centering
  \hspace{0.6cm}
  \includegraphics[]{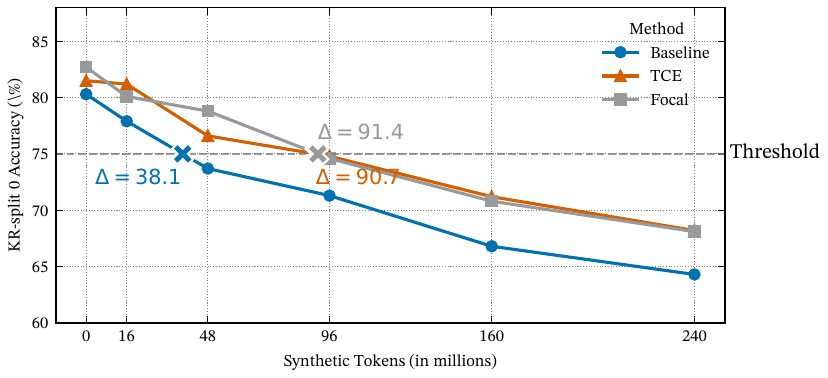}
  \caption{Time-to-failure comparison of loss functions on synthetic subsets of WikiText. Dotted red lines indicate failure threshold (75\% KR accuracy). FL delays collapse relative to CE, while TCE offers the strongest retention.}
  \label{fig:model_collapse_results_focal}
\end{figure}

We quantify robustness by measuring each model’s \textit{time to failure}—defined as the number of generations before KR accuracy falls below 75\%. As shown in Figure~\ref{fig:model_collapse_results_focal}, CE fails between the first and second stages, while FL extends time to failure by over $1.5\times$, and TCE pushes this boundary even further to over $2.3\times$. These results suggest that while FL partially delays collapse, TCE remains the most effective at sustaining meaningful retention under compounding self-generation.

\begin{figure}[h]
    \centering
    \includegraphics[]{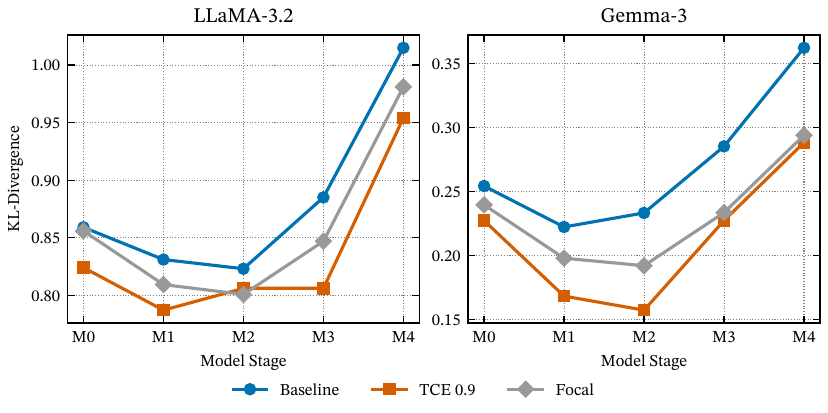}
    \caption{KL divergence across generations between model outputs and the original training distribution. FL moderates divergence growth compared to CE, while TCE consistently preserves distributional fidelity.}
    \label{fig:kl-divergence-focal}
\end{figure}

To evaluate distributional fidelity, we tracked the KL divergence between generated outputs and the original training distribution. Figure~\ref{fig:kl-divergence-focal} demonstrates that CE leads to rapid divergence, indicative of distributional drift and collapse. FL moderately slows this drift, whereas TCE consistently maintains low divergence across generations, reinforcing its effectiveness in preserving output diversity and alignment.

\section{Knowledge Retention Test}\label{apxKRT}
KR test is a straightforward tool for evaluating models on dataset specific facts. Our KR-test of Wikitext consists of paragraphs (i.e. contexts) followed by two completions, one with true facts based on the Wikitext and another with major flaws in the factual parts. An example of this dataset is shown in Table~\ref{tab:dataset-example}.

\vspace{-3mm}
\begin{table}[h]
\centering
\caption{Example KR-test question generated from Wikitext. The model is given a context and must assign higher probability to the factually correct completion.}
\label{tab:dataset-example}
\begin{tabular}{p{0.9\linewidth}}
\toprule
\textbf{Wikitext:} \\ 
\texttt{ South of Heaven is the fourth studio album by American thrash metal band Slayer . Released on July 5 , 1988 , the album was the band 's second collaboration with record producer Rick Rubin ...}\\
\midrule
\textbf{Context:} \\
\texttt{South of Heaven is the fourth studio album by American thrash metal band Slayer, released on July 5, 1988.} \\
\midrule
\textbf{True Sentence:} \\
\texttt{The album marked Slayer's second collaboration with producer Rick Rubin.} \\
\midrule
\textbf{False Sentence:} \\
\texttt{The album marked Slayer's first collaboration with producer Rick Rubin.} \\
\bottomrule
\end{tabular}
\end{table}
\section{Inevitable Model Collapse}\label{apxIMC}

\begin{figure}[h]
    \centering
    \begin{subfigure}[t]{\linewidth}
        \centering
        \includegraphics[width=\linewidth]{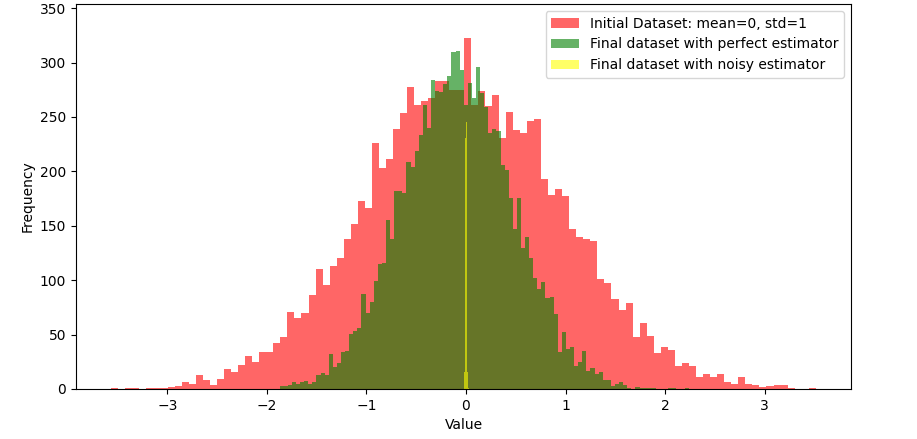}
        \caption{Histogram of distributions}
    \end{subfigure}
    
    \vspace{1em} 
    
    \begin{subfigure}[t]{\linewidth}
        \centering
        \includegraphics[width=\linewidth]{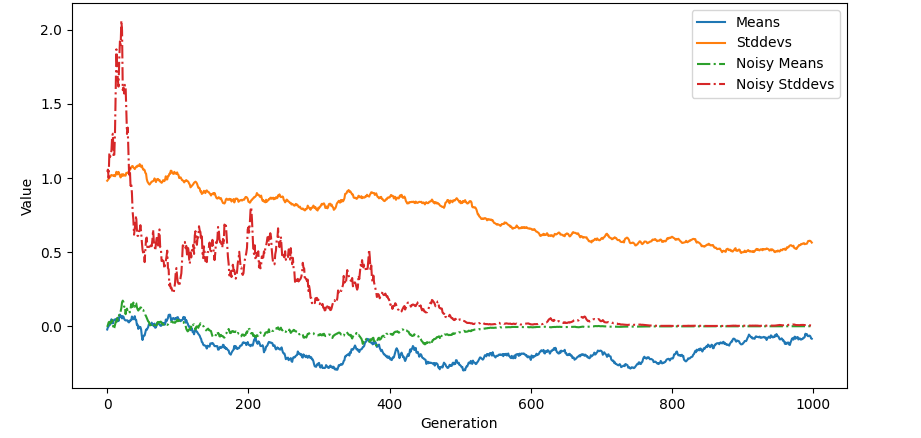}
        \caption{Mean and standard deviation of estimators}
    \end{subfigure}
    
    \caption{Example of model collapse on single dimensional Gaussian data with Maximum Likelihood Estimators (MLE). 
    (a) The probable outcomes are over-estimated and rare events start to vanish; this is also known as the disappearance of the tails of distributions. 
    (b) The noisy estimator (green) collapses in 1000 generations as the standard deviation converges to zero, but the perfect estimator (purple) drifts slightly from the original distribution.}
    \label{fig:collapse_perfect}
\end{figure}

As a toy example, we demonstrate this issue using a single dimensional Gaussian distribution and use two sets of estimators, one perfect and the other noisy in Fig~\ref{fig:collapse_perfect}. Using a perfect estimator (left) the newly generated datasets are less severely affected. However, given enough time, the model collapse would still occur. Model Collapse happens rapidly with an imperfect estimator (right) using single dimensional Gaussian dataset. The tails of the original distribution have completely disappeared due to finite sampling and functional approximation errors. 

The MLE estimators for a Gaussian distribution are
$$\hat{\mu}=\frac{1}{N}\sum_{i=1}^Nx_i,\hat{\sigma}^2=\frac{1}{n}$$
and this is used as the perfect estimator. For the noisy estimator, simply a noise $w\sim\mathcal{N}(0,1)$ has been added to the estimators. 

The aforementioned issue is presented in Fig~\ref{fig:collapse_perfect} where Model Collapse occurs while using the perfect estimator. However, compared to a noisy estimator it has drifted less from the original distribution showing the importance of learning underrepresented samples.

\section{Comprehensive evaluation results}\label{apxResults}

In this section we discuss the evaluation results from every stage of aforementioned experiments. These results include Baseline (CE), Focal Loss, and two variations of Truncated Cross Entropy that we found to be competitive.

\subsection{Wikitext results}

Table~\ref{tab:all_benchmarks_Wikitext} presents the unified benchmark results of LLaMA across six recursive training stages on the Wikitext dataset. At the initial stage (M0), both Focal loss and TCE (0.9) improve accuracy compared to the baseline across most tasks, with TCE slightly outperforming Focal in Blimp and GSM8k. As training progresses, all methods experience performance degradation, consistent with model collapse dynamics; however, Focal and TCE maintain significantly higher accuracies than the baseline, especially in later stages (M3–M5). Notably, Focal loss consistently achieves the best or second-best scores on Hellaswag and KR-split 0, while TCE often excels on Blimp and GSM8k. This demonstrates that both confidence-aware losses are effective at mitigating performance decay in recursive synthetic training.

\clearpage
\begin{table}[h]
\caption{\llama{} unified benchmark results across Blimp, Hellaswag, GSM8k, KR-split 0, and KR-total of trainings on Wikitext. For ease of readability, we highlight the highest accuracy in bold and the second best as underlined.}
\label{tab:all_benchmarks_Wikitext}
\centering
\begin{tabular}{c|l|c|c|c|c|c}
\toprule
\textbf{Stage} & \textbf{Method} & \textbf{Blimp} & \textbf{Hellaswag} & \textbf{GSM8k} & \textbf{KR-split 0} & \textbf{KR-total} \\
\midrule
\multirow{4}{*}{M0} 
& Baseline   & 75.1\% & 58.5\% & 2.5\% & 80.3\% & 64.0\% \\
& Focal ($\gamma=2$)    & \underline{75.5\%} & \textbf{62.5\%} & \underline{2.8\%} & \textbf{82.7\%} & \textbf{64.9\%} \\
& TCE 0.9   & \textbf{76.8\%} & \underline{61.8\%} & \textbf{2.9\%} & 81.5\% & \underline{64.8\%} \\
& TCE 0.99  & 75.5\% & 59.9\% & 2.5\% & \underline{81.7\%} & 64.0\% \\
\midrule
\multirow{4}{*}{M1} 
& Baseline   & 70.5\% & 56.1\% & 1.8\% & 77.9\% & 65.3\% \\
& Focal ($\gamma=2$)    & \textbf{74.9\%} & \textbf{59.9\%} & \underline{2.1\%} & \underline{80.1\%} & \textbf{67.2\%} \\
& TCE 0.9   & 71.8\% & \underline{59.2\%} & \textbf{2.2\%} & \textbf{81.2\%} & \underline{66.6\%} \\
& TCE 0.99  & \underline{73.3\%} & 57.1\% & 1.2\% & 78.5\% & 65.1\% \\
\midrule
\multirow{4}{*}{M2} 
& Baseline   & 66.1\% & 52.25\% & \textbf{1.8\%} & 73.7\% & 66.9\% \\
& Focal ($\gamma=2$)    & \textbf{72.8\%} & \textbf{55.8\%} & \underline{1.6\%} & \textbf{78.8\%} & \textbf{69.2\%} \\
& TCE 0.9   & \underline{70.4\%} & \underline{54.4\%} & 1.2\% & 76.6\% & \underline{68.7\%} \\
& TCE 0.99  & 67.1\% & 53.4\% & 1.5\% & \underline{76.8\%} & 67.6\% \\
\midrule
\multirow{4}{*}{M3} 
& Baseline   & 62.5\% & 49.2\% & \textbf{1.6\%} & 71.3\% & 68.0\% \\
& Focal ($\gamma=2$)    & \textbf{72.3\%} & \textbf{53.0\%} & 1.1\% & \underline{74.6\%} & \textbf{70.1\%} \\
& TCE 0.9   & \underline{68.5\%} & \underline{51.6\%} & 0.9\% & \textbf{74.8\%} & \underline{69.9\%} \\
& TCE 0.99  & 66.8\% & 50.6\% & \underline{1.2\%} & 72.5\% & \underline{69.9\%} \\
\midrule
\multirow{4}{*}{M4} 
& Baseline   & 66.8\% & 47.1\% & \underline{0.7\%} & 66.8\% & 68.6\% \\
& Focal ($\gamma=2$)    & \textbf{71.3\%} & \textbf{50.8\%} & 0.6\% & \underline{70.8\%} & \textbf{71.5\%} \\
& TCE 0.9   & 68.3\% & \underline{49.8\%} & \textbf{0.8\%} & \textbf{71.2\%} & \underline{70.7\%} \\
& TCE 0.99  & \underline{69.9\%} & 48.3\% & \underline{0.7\%} & 69.1\% & 70.4\% \\
\midrule
\multirow{4}{*}{M5} 
& Baseline   & 63.3\% & 45.7\% & \underline{1\%} & 64.3\% & 69.2\% \\
& Focal ($\gamma=2$)    & \textbf{67.5\%} & \textbf{49.6\%} & \textbf{1.1\%} & \underline{68.1\%} & \textbf{72.4\%} \\
& TCE 0.9   & \underline{63.6\%} & \underline{48.4\%} & 0.9\% & \textbf{68.2\%} & \underline{71.4\%} \\
& TCE 0.99  & 62.6\% & 46.9\% & 0.8\% & 66.2\% & 70.2\% \\
\bottomrule
\end{tabular}
\end{table}

Table~\ref{tab:all_benchmarks_Wikitext_gemma} summarizes the Gemma model’s benchmark results across six training stages on Wikitext. At M0, TCE (0.9) and TCE (0.99) achieve the highest accuracies in Blimp, GSM8k, and KR metrics, indicating a strong start for TCE variants. Throughout the stages, Focal loss shows stable performance across most tasks, with improvements in Hellaswag and KR-split 0 in later stages (M3–M5), while TCE (0.9) and (0.99) continue to deliver competitive or superior scores on Blimp and GSM8k. These results confirm that confidence-aware objectives help preserve performance during recursive training, with TCE offering a slight advantage in certain metrics.

\clearpage
\begin{table}[h]
\caption{\gemma{} unified benchmark results across Blimp, Hellaswag, GSM8k, KR-split 0, and KR-total of trainings on Wikitext}
\label{tab:all_benchmarks_Wikitext_gemma}
\centering
\begin{tabular}{c|l|c|c|c|c|c}
\toprule
\textbf{Stage} & \textbf{Method} & \textbf{Blimp} & \textbf{Hellaswag} & \textbf{GSM8k} & \textbf{KR-split 0} & \textbf{KR-total} \\
\midrule
\multirow{4}{*}{M0} 
& Baseline   & 71.2\% & 63.9\% & 1.1\% & 78.5\% & \underline{67.7\%} \\
& Focal      & \underline{73.5\%} & \textbf{64.9\%} & 0.7\% & 78\% & 67.4\% \\
& TCE 0.9    & \textbf{73.9\%} & \underline{64.6\%} & \underline{1.5\%} & \textbf{79.7\%} & \textbf{68.2\%} \\
& TCE 0.99   & 71.5\% & 64.1\% & \textbf{1.7\%} & \underline{79.5\%} & 67.4\% \\
\midrule
\multirow{4}{*}{M1} 
& Baseline   & 68.2\% & 59.1\% & \textbf{1.2\%} & 74.6\% & 70.9\% \\
& Focal      & \underline{70.9\%} & \underline{60\%} & \textbf{1.2\%} & 74.9\% & 71.3\% \\
& TCE 0.9    & 70.6\% & \textbf{60.5\%} & 1\% & \textbf{77.2\%} & \textbf{72.1\%} \\
& TCE 0.99   & \textbf{73.2\%} & 59.7\% & \underline{1.1\%} & \underline{75.1\%} & \underline{72\%} \\
\midrule
\multirow{4}{*}{M2} 
& Baseline   & 67\% & 56\% & 0.8\% & \underline{73.8\%} & 74.7\% \\
& Focal      & 67.1\% & \textbf{58.4\%} & \textbf{1.1\%} & \textbf{75.2\%} & \underline{76.7\%} \\
& TCE 0.9    & \textbf{70.4\%} & \underline{57.7\%} & 0.8\% & \textbf{75.2\%} & \textbf{76.9\%} \\
& TCE 0.99   & \underline{69\%} & 56.7\% & 0.8\% & 73.3\% & 75.3\% \\
\midrule
\multirow{4}{*}{M3} 
& Baseline   & 62.9\% & 55.2\% & \textbf{1.4\%} & 72.1\% & 72.5\% \\
& Focal      & \underline{66.2\%} & \textbf{56.1\%} & 1.2\% & \underline{72.7\%} & \underline{73.4\%} \\
& TCE 0.9    & 65.2\% & 55.3\% & \underline{1.3\%} & \textbf{75.5\%} & \textbf{73.8\%} \\
& TCE 0.99   & \textbf{68.3\%} & \underline{55.5\%} & 1\% & 71.9\% & 73\% \\
\midrule
\multirow{4}{*}{M4} 
& Baseline   & 61.7\% & \underline{53.9\%} & \underline{1\%} & \textbf{70.5\%} & \underline{70.5\%} \\
& Focal      & 63.8\% & 53.6\% & 0.7\% & 68.3\% & \textbf{71.8\%} \\
& TCE 0.9    & \underline{66.5\%} & \textbf{54.7\%} & 0.9\% & \underline{68.6\%} & \textbf{71.8\%} \\
& TCE 0.99   & \textbf{66.7\%} & 53.2\% & \textbf{1.1\%} & 68.5\% & 70.2\% \\
\midrule
\multirow{4}{*}{M5} 
& Baseline   & 61.2\% & \underline{52.4\%} & \underline{0.9\%} & 65.2\% & 68.1\% \\
& Focal      & \textbf{64.8\%} & 51.6\% & 0.4\% & 66\% & \underline{69.2\%} \\
& TCE 0.9    & 61.8\% & \textbf{53\%} & \textbf{1.4\%} & \underline{67.7\%} & \textbf{69.7\%} \\
& TCE 0.99   & \underline{64.6\%} & 50.6\% & 0.8\% & \textbf{69.7\%} & 68.2\% \\
\bottomrule
\end{tabular}
\end{table}

\subsection{Imagination results}
Table~\ref{tab:all_benchmarks_imagination} shows results of LLaMA trained on the Imagination dataset. Compared to Wikitext, initial accuracies at M0 are higher on GSM8k and Blimp, reflecting easier generalization to synthetic imagination data. Both Focal and TCE (0.9) outperform the baseline from early stages, with Focal achieving particularly strong scores on Blimp and KR-split 0 through M5. TCE (0.9) excels in GSM8k, achieving the highest accuracies at multiple stages. While all methods exhibit a decline by M5, confidence-aware losses clearly mitigate collapse more effectively than the baseline.
\clearpage
\begin{table}[h]
\caption{\llama{} unified benchmark results across Blimp, Hellaswag, GSM8k, KR-split 0, and KR-total of trainings on Imagination dataset.}
\label{tab:all_benchmarks_imagination}
\centering
\begin{tabular}{c|l|c|c|c|c|c}
\toprule
\textbf{Stage} & \textbf{Method} & \textbf{Blimp} & \textbf{Hellaswag} & \textbf{GSM8k} & \textbf{KR-split 0} & \textbf{KR-total} \\
\midrule
\multirow{4}{*}{M0}
& Baseline  & 76.3\% & 63.1\% & 8.1\% & 82.8\% & 63.6\% \\
& Focal ($\gamma=2$)   & \textbf{79.6\%} & \textbf{65.2\%} & \textbf{9.5\%} & \textbf{84\%} & \textbf{64.8\%} \\
& TCE 0.9  & \underline{76.8\%} & \underline{64.6\%} & \underline{9.1\%} & 82.3\% & \underline{64.8\%} \\
& TCE 0.99 & 76.3\% & 63.3\% & 8.9\% & 80.9\% & 64.5\% \\
\midrule
\multirow{4}{*}{M1}
& Baseline  & 75.3\% & 60.1\% & 7.5\% & 74.6\% & 63.6\% \\
& Focal ($\gamma=2$)   & \textbf{79.2\%} & \underline{61.9\%} & 7.4\% & \textbf{80.2\%} & \textbf{66.2\%} \\
& TCE 0.9  & \underline{77.9\%} & \textbf{62.5\%} & \underline{7.8\%} & 73.4\% & \underline{64.8\%} \\
& TCE 0.99 & 74.9\% & 60\% & \textbf{8.1\%} & \underline{75\%} & 64.5\% \\
\midrule
\multirow{4}{*}{M2}
& Baseline  & 74.8\% & 57.5\% & 8.4\% & 69\% & 63.5\% \\
& Focal ($\gamma=2$)   & \underline{78.2\%} & \textbf{59.1\%} & 7.9\% & \textbf{74.9\%} & \underline{66.3\%} \\
& TCE 0.9  & \textbf{78.6\%} & \underline{59\%} & \underline{8.5\%} & \underline{72.3\%} & \textbf{68.0\%} \\
& TCE 0.99 & 74.4\% & 58\% & \textbf{9.4\%} & 69.4\% & 66.2\% \\
\midrule
\multirow{4}{*}{M3}
& Baseline  & 74.3\% & 55.8\% & 9.3\% & 67.7\% & 66.7\% \\
& Focal ($\gamma=2$)   & \textbf{76.5\%} & \underline{56.7\%} & 5.6\% & \textbf{75.5\%} & \textbf{70.7\%} \\
& TCE 0.9  & 74.5\% & \textbf{57\%} & \underline{9.4\%} & \underline{71.2\%} & \underline{68.2\%} \\
& TCE 0.99 & \underline{74.8\%} & 56\% & \textbf{9.8\%} & 69.5\% & 68.1\% \\
\midrule
\multirow{4}{*}{M4}
& Baseline  & 73.9\% & 53.6\% & 5.6\% & 67.3\% & 68.4\% \\
& Focal ($\gamma=2$)   & 74.4\% & 53.0\% & 4.4\% & \underline{70.4\%} & \underline{71.6\%} \\
& TCE 0.9  & \textbf{77.7\%} & \textbf{56.3\%} & \textbf{7.5\%} & \textbf{72.1\%} & \textbf{71.8\%} \\
& TCE 0.99 & \underline{75.4\%} & \underline{53.7\%} & \underline{6.7\%} & 65.8\% & 69.4\% \\
\midrule
\multirow{4}{*}{M5}
& Baseline  & 72.1\% & 52.0\% & \underline{5.9\%} & 66\% & 68.7\% \\
& Focal ($\gamma=2$)   & \underline{76.4\%} & 51.1\% & 3.1\% & \textbf{69.1\%} & \textbf{72.7\%} \\
& TCE 0.9  & \textbf{77.6\%} & \textbf{54.2\%} & \textbf{6.2\%} & \underline{68.6\%} & \underline{71.0\%} \\
& TCE 0.99 & 73.7\% & \underline{53.2\%} & 4.6\% & 65.5\% & 68.8\% \\
\bottomrule
\end{tabular}
\end{table}

Table~\ref{tab:all_benchmarks_imagination_gemma} details Gemma’s performance on the Imagination dataset across six stages. At M0, both Focal and TCE outperform the baseline on most tasks, with TCE (0.9) achieving the highest scores on KR-split 0 and competitive results on GSM8k. Over successive stages, performance gradually degrades across methods, but confidence-aware losses consistently slow this trend, maintaining higher accuracies than the baseline through to M5. Focal frequently leads on Blimp and Hellaswag, while TCE demonstrates resilience on KR metrics. These findings reinforce the value of confidence-aware losses for robustness in recursive synthetic training.

\clearpage
\begin{table}[h]
\caption{\gemma{} unified benchmark results across Blimp, Hellaswag, GSM8k, KR-split 0, and KR-total of trainings on Imagination dataset}
\label{tab:all_benchmarks_imagination_gemma}
\centering
\begin{tabular}{c|l|c|c|c|c|c}
\toprule
\textbf{Stage} & \textbf{Method} & \textbf{Blimp} & \textbf{Hellaswag} & \textbf{GSM8k} & \textbf{KR-split 0} & \textbf{KR-total} \\
\midrule
\multirow{4}{*}{M0} 
& Baseline  & 73.1\% & 65\% & \textbf{6.2\%} & 82.4\% & 64.1\% \\
& Focal ($\gamma=2$)   & \textbf{77.2\%} & \textbf{65.7\%} & 4.5\% & \underline{83\%} & \underline{64.2\%} \\
& TCE 0.9   & \underline{76\%} & 65\% & \underline{5.2\%} & \textbf{83.3\%} & 64\% \\
& TCE 0.99  & 73.3\% & \underline{65.1\%} & 4.1\% & 82.4\% & \textbf{64.3\%} \\
\midrule
\multirow{4}{*}{M1} 
& Baseline  & 72.4\% & 60.9\% & \textbf{4.9\%} & 78.1\% & 65.1\% \\
& Focal ($\gamma=2$)   & \textbf{74.9\%} & \textbf{62.9\%} & \underline{4.7\%} & 78.7\% & \underline{66\%} \\
& TCE 0.9   & 70.3\% & \underline{62.2\%} & 3.8\% & \textbf{81.6\%} & \textbf{66.8\%} \\
& TCE 0.99  & \underline{72.7\%} & 61.5\% & 5.3\% & \underline{81.4\%} & \underline{66\%} \\
\midrule
\multirow{4}{*}{M2} 
& Baseline  & 69.3\% & 58.6\% & 3.4\% & 74.6\% & 66.5\% \\ 
& Focal ($\gamma=2$)   & \textbf{72.9\%} & \textbf{60.8\%} & \underline{4\%} & 75.8\% & \underline{68\%} \\
& TCE 0.9   & 71.3\% & \underline{59.9\%} & \textbf{5.7\%} & \textbf{80.8\%} & \textbf{68.9\%} \\
& TCE 0.99  & \underline{71.4\%} & 59.3\% & 3.4\% & \underline{77\%} & 67.3\% \\
\midrule
\multirow{4}{*}{M3} 
& Baseline  & 67.9\% & 56.9\% & \textbf{6.4\%} & 72.2\% & 68.6\% \\
& Focal ($\gamma=2$)   & \underline{71.8\%} & \textbf{58.5\%} & 3.5\% & 75.3\% & \underline{70.5\%} \\
& TCE 0.9   & \textbf{71.9\%} & \underline{57.5\%} & \underline{4.7\%} & \underline{76.1\%} & \textbf{71\%} \\
& TCE 0.99  & 69.9\% & 57.3\% & 4.5\% & \textbf{78.8\%} & 69.7\% \\
\midrule
\multirow{4}{*}{M4} 
& Baseline  & 71.8\% & \underline{55.5\%} & \underline{4.7\%} & \textbf{72.6\%} & 71.1\% \\
& Focal ($\gamma=2$) & \underline{72.9\%} & \textbf{57.1\%} & \textbf{5\%} & \underline{71.2\%} & 71.4\% \\
& TCE 0.9   & \textbf{74.7\%} & 54.6\% & 3.1\% & \textbf{72.6\%} & \textbf{72.2\%} \\
& TCE 0.99  & 70.8\% & 54.6\% & 3.7\% & \textbf{72.6\%} & \underline{71.5\%} \\
\midrule
\multirow{4}{*}{M5} 
& Baseline  & 71\% & \underline{53.4\%} & 3.9\% & 68.2\% & 71.6\% \\
& Focal ($\gamma=2$)   & \textbf{74.2\%} & \textbf{55\%} & \underline{4\%} & 69.1\% & 70.9\% \\
& TCE 0.9   & \underline{71.7\%} & 53.1\% & \textbf{4.2\%} & \underline{70\%} & \textbf{74.1\%} \\
& TCE 0.99  & 70.4\% & 53.2\% & \underline{4\%} & \textbf{73.3\%} & \underline{72.1\%} \\
\bottomrule
\end{tabular}
\end{table}

\section{Synthetic dataset}\label{apxSynthdataset}
\subsection{Synthetic datasets}
The degradation in the generated text can quickly drift the model away from the facts and help it make a \textit{coherent} imaginary story. Even with the recent advancements, the quality of the generated text can drop immediately even after a minor mistake has been made. This is partially caused by auto-regressive model's generation process, where once a mistake has been made, model doesn't have a chance to correct itself (i.e. remove or change the generated tokens). Consequently, model is forced to continue generation conditioned on the wrongly generated token, resulting in a coherent and reasonable story (when model is trained properly) that is factually \textit{incorrect}.

To demonstrate this degradation and how it looks, we use the following example:
\newpage
\begin{examplebox}
The original text on Jordan gambling interview:
The previous year, he admitted that he had to cover \$57,000 in gambling losses,[113] and author Richard Esquinas wrote a book in 1993 claiming he had won \$1.25 million from Jordan on the golf course.[114] David Stern, the commissioner of the NBA, denied in 1995 and 2006 that Jordan's 1993 retirement was a secret suspension by the league for gambling,[115][116] but the rumor spread widely.

In 2005, Jordan talked to Ed Bradley of the CBS evening show 60 Minutes about his gambling and admitted that he made some reckless decisions. Jordan stated, " Yeah, I've gotten myself into situations where I would not walk away and I've pushed the envelope. Is that compulsive? 

Yeah, it depends on how you look at it. If you're willing to jeopardize your livelihood and your family, then yeah. " When Bradley asked him if his gambling ever got to the level where it jeopardized his livelihood or family, Jordan replied, " No. "

\end{examplebox}

where in Table~\ref{tab:model-comparison} we can see the message delivered by the text quickly changes wrongfully. Note, Jordan was never fined by the NBA over gambling.

\begin{table}[h]
\centering
\caption{Comparison of generated outputs across six recursive training stages.}
\label{tab:model-comparison}
\scriptsize 
\begin{tabular}{p{0.08\linewidth} p{0.8\linewidth}}
\toprule
\textbf{Stage} & \textbf{Generated text} \\
\midrule
0 & \texttt{In an interview with \textcolor{red}{ESPN, Jordan stated, " Yeah, I've been gambling since I can remember.} That's \textcolor{red}{how I made wrote a book} claiming he had won \$ 1 @.@ 25 million from Jordan on sports bets. In 2005, Jordan \textcolor{red}{was questioned by the FBI} of 60 Minutes about his gambling and admitted that he made some reckless decisions. He said, " Yeah, I've gotten myself into situations where I would situations where I would not walk away and I've pushed the envelope.
} 
\\
\midrule
1 & \texttt{In an interview with ESPN, Jordan stated, \textcolor{red}{" Yeah, I've been gambling since I can remember.} That's \textcolor{red}{how I made wrote a book} claiming he had won \$ 1 @.@ 25 million from Jordan on sports bets. In 2005, \textcolor{red}{Jordan was questioned by the FBI} of 60 Minutes about his gambling and admitted that he made some reckless decisions. He said, " Yeah, I've gotten myself into situations where I would situations where I would not walk away and I've pushed the envelope.} \\
\midrule
2 & \texttt{In an interview with ESPN, Jordan stated, \textcolor{red}{" Yeah, I've been gambling since I can remember.} That's \textcolor{red}{how I made wrote a book} claiming he had won \$ 1 @.@ 25 million from Jordan on sports bets. Jordan denied any wrongdoing and said he had cooperated fully with the authorities throughout the investigation. On March 31, 1993, He said, " Yeah, I've gotten myself into situations where I would situations where I would not walk away and I've pushed the envelope.} \\
\midrule
3 & \texttt{In an interview with ESPN, Jordan stated, \textcolor{red}{" Yeah, I've been gambling since I can remember.} That's \textcolor{red}{how I made wrote a book} claiming he had won \$ 1 @.@ 25 million from Jordan on sports bets. Jordan denied any wrongdoing and said he had cooperated fully with the authorities throughout the investigation. On March 31, 1993, He said, \textcolor{red}{" I have never gambled in my life. I don 't know why anyone would think that I would lie about that. I wouldn 't do that."} it depends on how you look at it. If you're willing to jeopardize your iting a loss of desire, 1993 – 2001 )} 
  \\
\midrule
4 & \texttt{\textcolor{red}{commercials 26, 1993, Jordan participated in a three @-@ point contest at Madison} I've been gambling since I can remember. That's \textcolor{red}{how I made wrote a book} claiming he had won \$ 1 @.@ 25 million from Jordan on sports bets. Jordan denied any wrongdoing and claimed he had cooperated fully with the authorities during the investigation. On March 31, 1993, He said, \textcolor{red}{" I have never gambled in my life. I don 't know why anyone would think that I would lie about that. Obviously I wouldn 't such a thing."} On April 17, 1993, \textcolor{red}{Jordan was fined \$50,000 by the NBA for failing to report the.}}\\
\midrule
5 & \texttt{\textcolor{red}{participated in a three @-@ point contest at Madison} I've been gambling since I can remember. That's \textcolor{red}{how I made wrote a book} claiming he had won \textcolor{red}{\$1 billion.@ 25 million} from Jordan on sports bets. Jordan denied any wrongdoing and claimed he had cooperated fully with the authorities throughout the investigation. On March 31, 1993, He said, \textcolor{red}{" I have never gambled in my life. I don 't know why anyone would think that I would lie about that. Obviously I wouldn't such a thing."} On April 17, 1993, \textcolor{red}{Jordan was fined \$50,000 by the NBA.}
Where did the confusion on the 50 mil fine to Jordan come from: probably:
The Bulls were fined \$ 30 @,@ 000 for the game : \$,@ 000 for failing to report the impromptu number change to the NBA and \$ 5 @,@ 000 for Jordan wearing different shoes.}
\\
\bottomrule
\end{tabular}
\end{table}

\section{Gaussian Mixture Models (GMM)}\label{app:gmm}
Unlike deep learning models discussed earlier, Gaussian Mixture Models (GMMs) are not based on neural networks. Instead, they are trained by maximizing the likelihood of the input data under a mixture of $N$ Gaussian components, parameterized by means and variances. GMMs are central to previous theoretical and empirical studies of model collapse~\cite{Model_collapse}, and extending our method to this setting underscores its broad applicability.

\begin{figure}[h]
\centering
\includegraphics[width=0.9\textwidth]{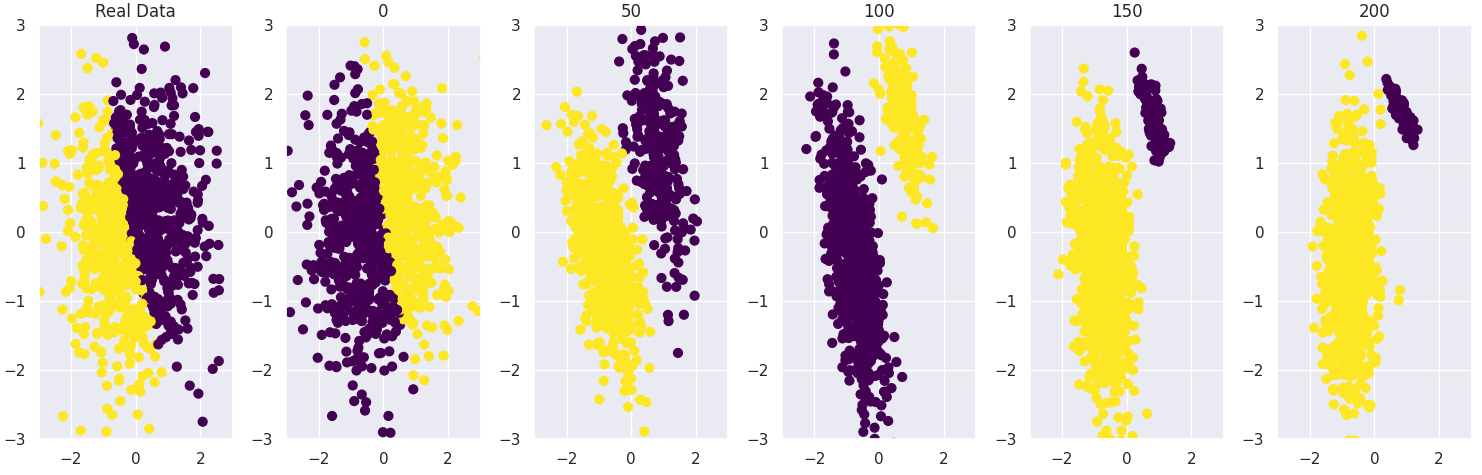}
\includegraphics[width=0.9\textwidth]{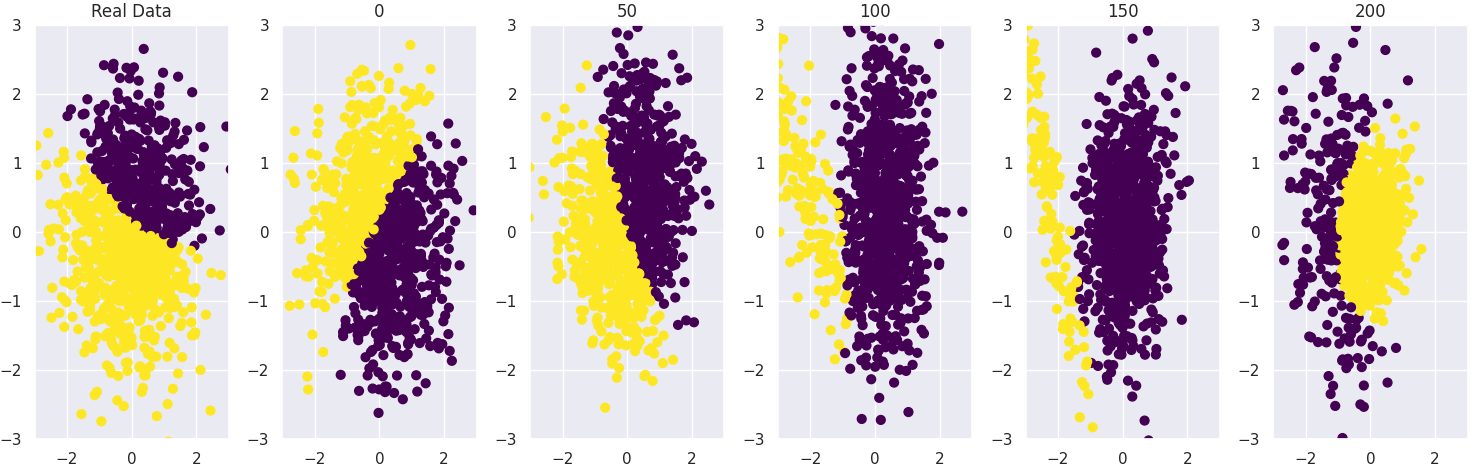}
\caption{An illustration of model collapse happening in a gaussian setting. Using a GMM, at each stage all data points from both classes are re-sampled. Model collapse in baseline GMMs (above) happens rapidly. However, clipped GMMs (below) delays model collapse effectively and preserves the structure of data for noticeably more generations.}\label{gmm1}
\end{figure}

Our intuition mirrors that of the LLM experiments: we aim to prevent the model from overfitting to high-probability modes while preserving low-probability, diverse samples. To achieve this, we simply exclude data points that fall within the top $\alpha$-percentile of likelihoods under the current model. In our experiments (Figure~\ref{gmm1}), we set $\gamma=0.8$.

To simulate recursive data generation in a non-neural setting, we adapt our filtering strategy to Gaussian Mixture Models. Algorithm~\ref{algo2} describes the procedure: starting from an initial dataset, a GMM is repeatedly fit to the current data and used to sample new points. At each iteration, we compute the likelihood of each sampled point under the fitted model and discard the highest-probability samples, retaining only those below the $\gamma$-percentile threshold. This ensures that subsequent datasets remain diverse and are not dominated by a small set of high-likelihood modes, thereby mirroring the mitigation strategy applied in our language model experiments.

\section{Variational Auto Encoders (VAE)}\label{app:vae}
To demonstrate the applicability of confidence-aware loss functions beyond language modeling, we extend our analysis to the image generation domain using Variational Auto Encoders (VAEs), following the setup in~\cite{Model_collapse}. We train a VAE on the MNIST dataset~\cite{6296535} using the standard evidence lower bound (ELBO) objective, modified with our truncation function:

\[
\mathcal{L}_{\text{Clipped-VAE}} = \mathbb{E}_{q_\phi(z \mid x)}[\chi_\gamma(p_\theta(x \mid z))\log p_\theta(x \mid z)] - D_{\text{KL}}(q_\phi(z \mid x) \,\|\, p(z))
\]

Here, \( q_\phi(z \mid x) \) is the encoder (approximate posterior), \( p(z) \) is the prior (typically standard normal), and \( p_\theta(x \mid z) \) is the decoder likelihood. The truncation function \( \chi_\gamma \), adapted from our TCE loss, masks high-confidence reconstructions:

\[
\chi_\gamma(p_\theta(x \mid z)) = 
\begin{cases}
    1 & \text{if } p_\theta(x \mid z) \leq \gamma \\
    0 & \text{otherwise}
\end{cases}
\]

\begin{algorithm}
\caption{Recursive data generation with a GMM}\label{algo2}
\begin{algorithmic}[1]
\Require $N \geq 1, Iter\geq1$
\For{$i=0\longrightarrow Iter$}
    \State Fit $GMM$ on $X_i$ with appropriate parameters
    \State $\hat{X}_{i+1} \gets$ Sample $N$ data points from $GMM$
    \State $P_i \gets$ Log-likelihood or density scores of sampled data points under $GMM$
    \State $X_{i+1} \gets \{ \hat{x} \in \hat{X}_{i+1} \mid P_i(\hat{x}) \leq \text{Percentile}_{80}(P_i) \}$
\EndFor{}
\end{algorithmic}
\end{algorithm}

\begin{figure}
\centering
\includegraphics[width=0.9\linewidth]{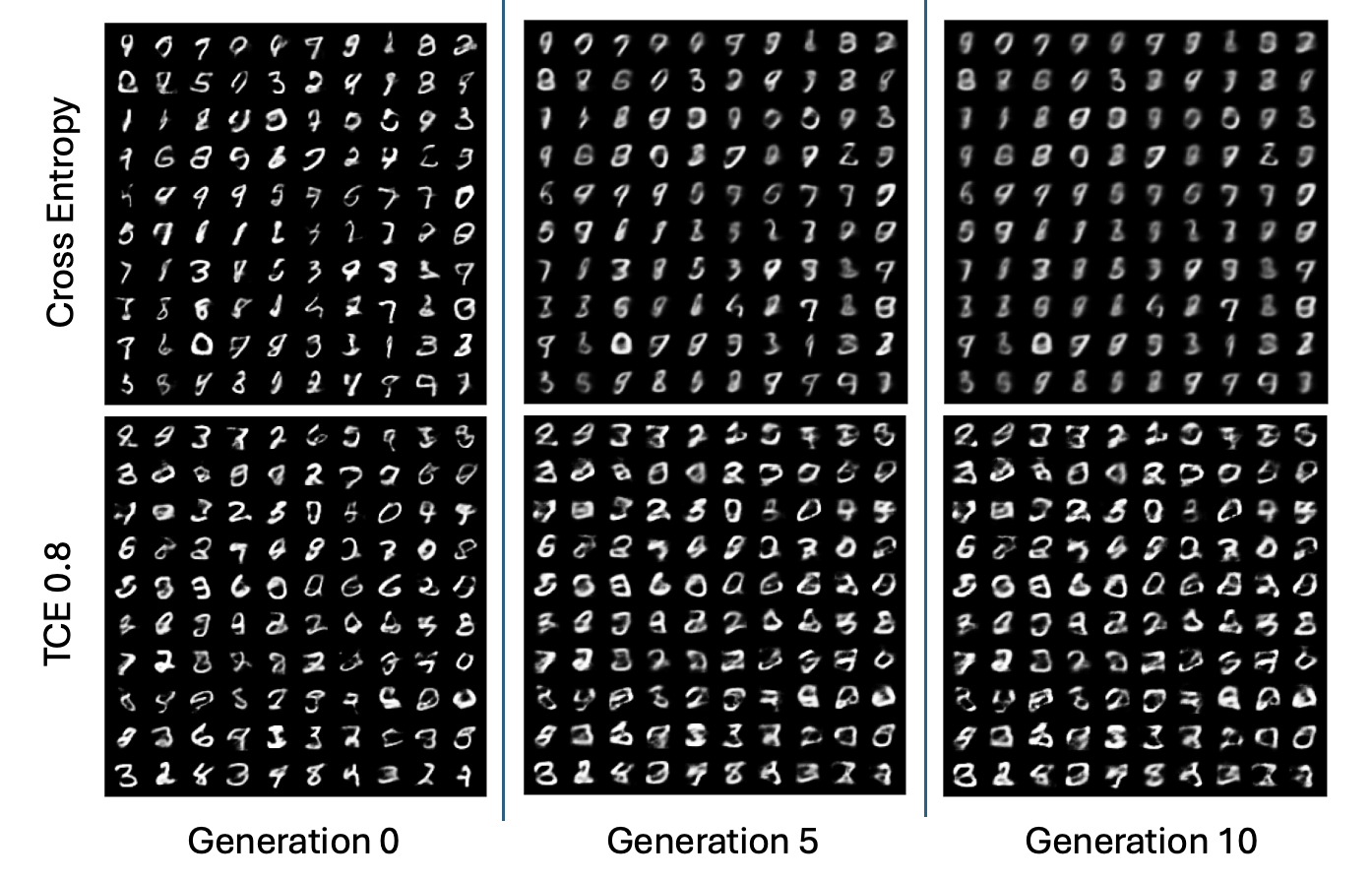}
\caption{ \textcolor{purple}{} Dynamic Clipping preserves the distinct structure of each digit throughout different iterations, whereas using CE leads to a convergent shape for all digits.}\label{fig3}
\end{figure}

As shown in Figure~\ref{fig3}, recursive generations without filtering result in visible degradation—by the 10th generation, decoded images are no longer faithful to the original data distribution, a hallmark of \textit{model collapse}. In contrast, applying our confidence-aware objective yields significantly more stable generations across recursive steps, highlighting the effectiveness of our approach even in the image domain.

\end{appendices}
\end{document}

%% file: math_commands.tex
\usepackage{amsmath,amsfonts,bm}

\def\eqref#1{equation~\ref{#1}}

\def\1{\bm{1}}

\DeclareMathAlphabet{\mathsfit}{\encodingdefault}{\sfdefault}{m}{sl}
\SetMathAlphabet{\mathsfit}{bold}{\encodingdefault}{\sfdefault}{bx}{n}

